\documentclass[11pt,a4paper]{article}
\usepackage{times}
\usepackage{latexsym}
\usepackage{arxiv}
\usepackage{hyperref}       

\usepackage{microtype}

\title{LUC at ComMA-2021 Shared Task: Multilingual Gender Biased and Communal Language Identification without using linguistic features 
}

\author{Rodrigo Cuéllar-Hidalgo \\
  CENIDET/COLMEX \\
  \texttt{\small  rcuellar@colmex.mx} \\
 
   \And
  Julio de Jesús Guerrero-Zambrano \\
  IIM \\
  \texttt{\small julio@jzambrano.xyz} \\  
  
  \AND
  Dominic Forest \\
  Université de Montréal \\
  EBSI \\
  \texttt{\small dominic.forest@umontreal.ca} \\  
  \And
  Gerardo Reyes-Salgado \\
  CENIDET \\
  \texttt{\small gerardo.rs@cenidet.tecnm.mx} \\  
\And
Juan-Manuel Torres-Moreno\\
  Université d'Avignon \\
  LIA \\
  \texttt{\small juan-manuel.torres@univ-avignon.fr} \\
  }

\date{}

\begin{document}
\maketitle
\begin{abstract}
This work aims to evaluate the ability that both probabilistic and state-of-the-art vector space modeling (VSM) methods provide
to well known machine learning algorithms to identify social network documents to be classified as aggressive, gender biased or communally charged. 
To this end, an exploratory stage was performed first in order to find relevant settings to test, i.e. by using training and development samples, we trained multiple algorithms using multiple vector space modeling and probabilistic methods and discarded the less informative configurations. These systems were submitted to the competition of the ComMA@ICON'21 Workshop on Multilingual Gender Biased and Communal Language Identification.
\end{abstract}

\section{Introduction}

The introduction of the Internet and its democratization in the public sphere has fostered the emergence of many sociological phenomena. This opens the possibility of forming friendly relations and information sharing from online networking platforms \cite{Arroyo18}. As the organizers say: \textsl{``Aggression and its manifestations in different forms have taken unprecedented proportions with the tremendous growth of Internet and social media.''} \footnote{\url{https://competitions.codalab.org/competitions/35482}}. 
The challenge ComMA@ICON 2021 is an interesting task in order to automatically discover and understand the pragmatic and structural aspects of such forms of language usage \cite{waseem,Dadvar}. 

Our international LUC$^2$ team\footnote{Team composed by LIA (Laboratoire Informatique d'Avignon/Université d'Avignon, France), EBSI/Université de Montréal (Québec, Canada), CENIDET (Centro Nacional de Investigación y Desarrollo Tecnológico, Cuernavaca, México) and COLMEX (the Colegio de México, CDMX, México).} have worked in such challenge using several approaches mainly without linguistic features.

This paper is organized as follows: the section \ref{art} presents some relevant state of the art work related to automatic aggression identification, section \ref{methodo} describes the methodology and the ComMA@ICON’21 dataset used, section \ref{results} shows the results and finally the section \ref{conclusion} analyzes and discusses the contributions of this paper.

\section{Related work}
\label{art}

There already exist software designed to identify aggression and cyberbullying in social medias, e.g. CyberPatrol. The main drawback of these systems is that they are based on
keyword filtering, which is a limitation because no statistical features of texts are taken into account.
Further, these keyword filtering methods require manual maintenance.
To overcome the limitations of keyword filtering systems, \cite{Yin} is one of the former attempts to detect cyberbullying by using statistical features: word frequency, analysis of
feelings
(use of
pronouns in the second person, insults, etc.) and context. \cite{Dinakar_Reichart_Lieberman_2021} built a system that can
detect bullying elements in commentaries of YouTube videos. These were classified according to different representative categories (sexuality, intelligence, race and physical attributes). The classification
revealed weaknesses and an increase in false positive cases. Researchers emphasized the importance
of using common sense to understand users’ goals, emotions, and relationships, thereby disambiguating
and contextualizing language.
In \cite{BerryKogan2010} the authors were also interested in a word search method based on a bag
of words (BoW) system incorporating sentiment and contextual analysis. They build a decision tree that
predicted intimidating messages with an accuracy rate of 0.93. The researchers also have developed the
Chatcoder software to detect malicious activities online \cite{Kontostathis2012IdentifyingPU}.

Another study tested a system that allows users of a website to control the messages posted
on their web pages: it customized vocabulary filtering criteria using a machine learning method that
automatically labeled the contents. This approach had limitations because it was unable to measure
relationships between terms beyond a certain semantic level \cite{af923d24e041439aab3a7fd01c3a7d55}.
\cite{journals/wias/NaharLZP14} provided a concrete method for detecting online harassment by measuring the
score of sent and received messages (and thus their degree of involvement in a conversation) using the
Hyper-link Induced Topic Research algorithm (HITS). The authors also proposed a graphical model that
identifies the aggressors and their most active victims.
Other studies have attempted to go further by seeking to take into account more specific characteristics.
\cite{af923d24e041439aab3a7fd01c3a7d55} tried to establish a system based on language features characterizing the author's genre of comments on MySpace. Their results revealed an improvement in the detection of bullying
when this information is taken into account.
As we can see, recent work defines the means to respond to the cyberbullying phenomenon that is becoming more and more widespread as the use of the web does.

The problem of identifying offensive and abusive language is a more difficult task than expected due to the prevalence of 5 factors, according to \cite{nobata2016abusive}, described below:

\begin{enumerate}
\item The intentional obfuscation of words that lead to false positives.

\item The difficulty in identifying racial slurs since depending on the target group, this can be offensive or flattering.

\item The grammatical fluidity that leads to false negatives.

\item The limit of sentences, that is, abusive language can be articulated in more than one sentence.

\item The sarcasm, which is even difficult for a human to interpret, implies a lot of knowledge about the context of the message (geographical, historical, social, etc).
    
\end{enumerate}

Other aspects detected by \cite{nobata2016abusive} corresponds to the heterogeneity in the approach to the problem itself that causes too much noise and confusion, such as the fact of only addressing specific aspects, specific contexts, different definitions for certain terms and / or domain, and finally different sets of assessment.

At present, the task of classifying text in "agglutinating" or "morphologically rich" languages, leaves aside classical preprocessing and is replaced by the use of deep neural networks and word embeddings, since they take into account the semantic distance of the words in context, which is very useful in categorization tasks, which is not the case with the classic bag of words methods. A clear example of this is the implementation of fastText for the Turkish language by \cite{Kuyumcu2019}.

 \section{Methodology and data-sets}
\label{methodo}

In this section we present the methodology and the data-sets used in our study.

\subsection{Data-sets}

For this task, the ComMA organizers have provided a multilingual data-set with a total of 12,000 instances for training and development and an overall 3,000 instances for testing. 
The corpus is in three Indian languages Meitei, Bangla (an Indian variety) and Hindi and English. Several instances are expressed in two our more languages.
Refer to the challenge website for further information\footnote{\url{https://competitions.codalab.org/competitions/35482\#learn_the_details-datasets}}. 

From the organizer's website, the corpus is labelled as follows:
\begin{enumerate}
    \item {\bf Aggression level}. This category gives a classification in ‘Overtly Aggressive’(OAG), ‘Covertly Aggressive’(CAG) or ‘Non-aggressive’(NAG) text.

    \item {\bf Gender}. This category classify the text as ‘gendered’(GEN) or ‘non-gendered’(NGEN).

    \item {\bf Communal}. The task is to develop a binary classifier for classifying the content as ‘communal' (COM) or 'non-communal'(NCOM).
\end{enumerate}

We confirmed that the data-set furnished (both training and validation) are too small in order to employ Deep Learning methods. 
We then chose to use mainly classical probabilistic and VSM \cite{salton1983extended} oriented algorithms.

\begin{table}[!]
\centering
\begin{tabular}{rrrr}\hline 
\bf Corpus & \bf Instances & \textbf{Tokens} & \textbf{Chars} \\ \hline
Training & 9,000 & 186,017 & 1 585,979 \\
Developing & 3,000 & 55,996 & 473,403\\ \hline
Testing & 3,000 &  82,367 & 815,104 \\
\hline
\end{tabular}
\caption{\label{table-dataset} Basic statistics from ComMA corpus.}
\end{table}

\subsection{Pre-processing}

Because there are a mixture of several languages (Meitei, Bangla, Hindi and English) and often the data-set instances presents two or more languages mixed, we decided of avoid any linguistic pre-processing.
Indeed, neither stemmer, filtering or lemmatizer was used in our study. The only preprocessing that was carried out was the removal of NaN  and extraction of basic characteristics of each message, which are listed below:
\begin{itemize}
\item Number of words.
\item Number of sentences.
\item Number of scores.
\item Number of numbers.
\item Number of unrecognizable characters (emojis).
\end{itemize}

Using a simple tokenizer written in Python.

\subsection{Algorithms}

To tackle the problem presented in this challenge, we develop LUC, a multi-classifier that uses the following algorithms:

\begin{itemize}
    \item Nearest-Neighbor algorithm (KNN) \cite{KNN};
    \item Naive Bayes method \cite{lewis1998naive};
    \item Support Vector Machine algorithm (SVM) \cite{cortes1995support};
    \item Random Forest algorithm \cite{breiman2001random};
    \item Generalized Boosted Regression Models (GBM) \footnote{https://github.com/gbm-developers/gbm};
    \item Adaboost \cite{freund1997ada};
    \item Neural Networks (NN) based algorithms \cite{Hopfield}.
\end{itemize}

In the first system (S1), the individual outputs of the algorithms Naive Bayes, SVM, Random Forest, GBM, Adaboost and a multi-layer perceptron were combined in a single output using a mixing strategy that assembles all of the models that were created using the previous algorithms. In order to combine the predictions of the previously mentioned estimators, it was necessary to stack the outputs of each individual estimator and use a final estimator to compute the final prediction.

The stacking classifier responsible to compute the final estimation takes every individual estimator as input and creates a final estimator by training cross-validated predictions of the base estimators. For each estimator, the final classifier computes the prediction probability and final prediction, to return a final estimation based on a logistic regression of the inputs.

In order to achieve better results, each one of the tasks were trained and executed independently (by language and category) and the results were combined at the end. Accuracy was measured in order to keep track of the metrics of the results.

In the second system (S2), we also explored the relevance of the K nearest neighbors (KNN) algorithm alone to perform these supervised classification tasks. This algorithm is well known in the field of machine learning. It is both simple and efficient. It consists in assigning to each document of the test set the category with the highest frequency among its K nearest neighbors. The cosine measurement was used to evaluate the distance between the vectors representing each document. In addition, we varied two main parameters during the learning phase. On the one hand, we have varied the value of K, that is to say the number of neighbors to be considered. We systematically compared the results using 1, 2, 3, 4, 5, 10, 15, 20, 25 and 50 neighbors.  We also varied the number of features to describe the documents. As mentioned previously, no preprocessing was applied to reduce the size of the initial lexicon. Based on the frequency of strings in the entire corpus and by evaluating the correlation between the most frequent strings and the categories to predict (using a simple Chi2 measure), we used from 500 to 30,000 strings of characters (in increments of 500) to describe the training corpus. During the learning phase, we obtained the best results using 30,000 features and K = 1. It is therefore this combination that we applied to the test corpus (system 5).

\section{Results}
\label{results}

We present our results in two parts. 
In the first one we show the accuracy of methods on the training set. In the second one, the official F1 measures from the CodaLab website of the organizers are showed in their ``Results'' page\footnote{See: \url{https://competitions.codalab.org/competitions/35482\#results}}.

\subsection{Training and Developing data-sets}
\label{sect:pdf}

For the training set, our main results are showed in the Table~\ref{table-train}.

In the training phase, the accuracy was our principal criterion to optimize on the Developing data-set.

\begin{table}[!]
\centering
\begin{tabular}{crr}
\hline  & \textbf{KNN (S2)} & \textbf{Combined (S1)} \\ \hline
Accuracy & 0.98 & 0.7746 \\
\hline
\end{tabular}
\caption{\label{table-train} Developing set evaluation of our best classifiers. }
\end{table}

\subsection{Testing set}
\label{ssec:testing}

The main results on the testing set are shown in the Table~\ref{test-table}.
It is possible to see that our classifier LUC obtains the 3rd rank on the multilingual task, with a overall-F1 measure of {\bf 0.234} (see Addendum).

\begin{table}[!]
\centering
\begin{tabular}{crr}
\hline \textbf{F1} & \textbf{KNN (S2)} & \textbf{Combined (S1)} \\ \hline
Instance F1 & 0.234 & 0.213 \\
Overall micro-F1 & 0.615 & 0.616 \\
A. micro-F1 & 0.446 & 0.389 \\
G. bias micro-F1 & 0.675 & 0.693 \\
C. bias micro-F1 & 0.726 & 0.766 \\
\hline
\end{tabular}
\caption{\label{test-table} Testing set evaluation for our classifiers. {See: \url{https://competitions.codalab.org/competitions/public_submissions/35482}}.}
\end{table}


\section{Discussion and Conclusions}
\label{conclusion}

This task was very challenging.
The corpus (train, developing and test) are very noisy at several levels (syntactic, lexical, semantic, etc.),  and the instances presents two or more languages mixed. There are instances that make no sense. By example, the instances from train and developing sets:

\begin{verbatim}
C575.31	rrrrrrrrrrrrrrrrrrrrrrrrrr
rrrrrrrrrrrrrrrrrrrrrrrrrrrrrrurrr
rrrrrrrrrrrrryrrrr (NAG,NGEN,NCOM)

C575.690.1  XXXXXXXXXXXXXXXXXXXXXX
(NAG,NGEN,NCOM)

C579.732.1	G	(NAG,NGEN,COM)
\end{verbatim}

\noindent where the string of ``r'' of the instance C575.31 represents a string of a dog's emojis, is a good example of semantic noise.
What was the logic (if any) of the annotators in assigning the classes (NAG,NGEN,NCOM) to this instance? 


However, our strategy has obtained the third rank on overall-F1 measure and the second rank on the Aggression micro-F1 measure.
The methodology that we adopted in this study allowed us to observe in broad strokes the complexity of the challenge.

We think that our model may be enriched by other model sentences representation, as followed in \cite{ARROYOFERNANDEZ}, in order to outperform our results.
Finally, sentences classified with labels 'aggression', 'genre' or 'communal levels' may be used in other NLP tasks, by example on the summaries generation guided by a specific context and coming from social network text data \cite{torres}.

\section*{Acknowledgments}

This work was partially financed by the COLMEX (Mexico), CENIDET (Mexico), Instituto de Ingenieros de Morelos (Mexico), LIA (Université d'Avignon, France) 
and by the EBSI (Université de Montréal, Québec). 

\section*{Addemdum}

We dispute the results mentioned on the challenge's website. Indeed, the organizers of the challenge allowed certain teams to benefit from an additional day for the submission of their result. This violates the original rules of the challenge and has favored some teams. Strangely, the teams that benefited from this advantage finished in the top 3 positions. We therefore consider, by virtue of the initial regulations that we respected, that we finished in 3rd place in the task on the multilingual corpus. We are also very amazed at how the results have evolved over time. The results were first published online on the challenge website (we were then 3rd), then these results were modified (we were then ranked 5th). Finally, the non-public excel sheet was sent to the teams with new results, without any explanation (we ended up in 6th place).

\bibliography{biblio}
\bibliographystyle{plain}



\end{document}